\pdfoutput=1

\documentclass[11pt]{article}

\usepackage{acl}

\usepackage{times}
\usepackage{latexsym}

\usepackage[T1]{fontenc}

\usepackage[utf8]{inputenc}

\usepackage{microtype}

\usepackage{inconsolata}

\usepackage{graphicx}

\usepackage{booktabs}
\usepackage{xspace}
\usepackage{makecell}

\newcommand{\situa}[0]{S8D\xspace}
\newcommand{\disto}[0]{Dist\xspace}

\newcommand{\resili}[0]{ReLM\xspace}
\newcommand{\plt}[0]{PLT\xspace}
\newcommand{\hart}[0]{HaRT\xspace}

\newcommand{\hartwbft}[0]{{HaRT\tiny{WB-FT}}\xspace}
\newcommand{\hartwbftridge}[0]{{\hartwbft\tiny{+ Ridge}}\xspace}
\newcommand{\hartadapft}[0]{{HaRT\tiny{adaptive-FT}}\xspace}
\newcommand{\hartmaladapft}[0]{{HaRT\tiny{maladaptive-FT}}\xspace}

\usepackage{natbib}

\title{Who We Are, Where We Are: Mental Health at the Intersection of Person, Situation, and Large Language Models}

\author{Nikita Soni$^1$, August Håkan Nilsson$^2$, Syeda Mahwish$^1$, \\ \textbf{Vasudha Varadarajan$^1$, H. Andrew Schwartz$^1$, Ryan L. Boyd$^3$} \\
$^1$Department of Computer Science, Stony Brook University \\
$^2$Oslo Metropolitan University, Oslo Business School \\
$^3$University of Texas at Dallas, Dept. of Psychology \\
\texttt{\{nisoni, smahwish, vvaradarajan, has\}@cs.stonybrook.edu}, \\ august.nilsson1907@gmail.com, boyd@utdallas.edu}

\begin{document}
\maketitle
\begin{abstract}
Mental health is not a fixed trait but a dynamic process shaped by the interplay between individual dispositions and situational contexts. Building on interactionist and constructionist psychological theories, we develop interpretable models to predict well-being and identify adaptive and maladaptive self-states in longitudinal social media data. Our approach integrates person-level psychological traits (e.g., resilience, cognitive distortions, implicit motives) with language-inferred situational features derived from the Situational 8 DIAMONDS framework. We compare these theory-grounded features to embeddings from a psychometrically-informed language model that captures temporal and individual-specific patterns. Results show that our principled, theory-driven features provide competitive performance while offering greater interpretability. Qualitative analyses further highlight the psychological coherence of features most predictive of well-being. These findings underscore the value of integrating computational modeling with psychological theory to assess dynamic mental states in contextually sensitive and human-understandable ways.
\end{abstract}

\section{Introduction}

Understanding mental health through language has long been a foundational goal in clinical psychology and computational social science. Human expression — especially as manifested through digital communication — provides a unique window into internal states, social interactions, and psychological well-being. The CLPsych 2025 shared task builds on prior work in computational linguistics and clinical psychology, extending the analysis of mental health from static assessments to a dynamic, temporally anchored perspective. Seq2Psych, our interdisciplinary team, approaches this challenge by integrating psychological theory with computational methods, ensuring that our models are both empirically grounded and practically applicable. Our work emphasizes not only predictive accuracy but also a principled, theory-grounded approach to interpretability, facilitating a nuanced explanation of how self-states fluctuate over time.

\textbf{Primary Contributions} made in this work include: (1) proposal of a theory-driven baseline that combines language-inferred person-level traits (e.g., well-being, cognitive distortions, resilience) with situational context features derived from the Situational 8 DIAMONDS framework~\cite{rauthmann_situational_2014}; (2) use of a human-centered language model~\cite{soni-etal-2022-human, soni-etal-2024-large} trained on temporal user histories, to generate person-contextualized embeddings aligned with psychometric theory; (3) evaluation of these representations — individually and in hybrid configurations — for predicting well-being and identifying adaptive/maladaptive self-states in longitudinal text data; (4) an analysis of the most predictive psychological features to highlight interpretable connections between language, context, and mental health outcomes.




\section{Background}

Traditional models of psychological assessment often rely on static categories — diagnostic labels that imply stable traits or enduring conditions. However, integrative psychological theories emphasize that mental states are inherently dynamic, shaped by a complex interplay between individual dispositions and situational contexts \cite{buss_selection_1987, ekehammar_interactionism_1974, boyd_verbal_2024}. The constructionist view of emotions, for example, posits that emotional experiences emerge from interactions between an individual’s traits, cognitive processes, and environmental affordances \cite{barrett_how_2017}. Likewise, interactionist approaches in personality psychology highlight that adaptiveness or maladaptiveness of a given behavior is highly contingent upon situational fit \cite{mischel_cognitive-affective_1995, fleeson_moving_2004}.

The notion of situational fit is central to understanding mental health dynamics. Psychological well-being is not merely an individual trait but a function of how well a person's responses align with their context. A behavior that is adaptive in one situation may be maladaptive in another. For instance, hypervigilant behaviors may be adaptive in some contexts (e.g., military personnel in combat situations), but highly \textit{maladaptive} in another (e.g., a classroom or casual social gathering; \citealt{vyas_cognitive_2023}). This perspective aligns with the broader movement in psychology that views well-being as a dynamic process rather than a fixed state \cite[see, e.g.,][]{hollenstein_this_2015}

\subsection{A Principled Baseline: Integration of Person-Level Traits and Situational Context}
\label{sec: princi_base}

To model mental health dynamics in a principled manner, our approach combines person-level traits with psychological dimensions of the situation. Specifically, we leverage:

\paragraph{Psychological Characteristics of the Situation} Using a large language model, we annotated each post for the psychological characteristics of its context, based on the Situational 8 DIAMONDS (S8D) framework \cite{rauthmann_situational_2014}. This framework captures eight psychosocial aspects of a given situation — \textbf{D}uty, \textbf{I}ntellect, \textbf{A}dversity, \textbf{M}ating, p\textbf{O}sitivity, \textbf{N}egativity, \textbf{D}eception, and \textbf{S}ociality — that shape the meaning and context of person-environment transactions.

\paragraph{Person-Level Traits (PLT)} We employed existing models and methods to assess and estimate individual differences in implicit motives \cite{nilsson2025automatic}, depression and anxiety \cite{gu_natural_nodate}, harmony in life and satisfaction with life \cite{kjell_natural_2022} valence \cite{eijsbroek_comparison_2023}, cognitive distortions ~\cite{varadarajan-etal-2025-linking}, and resilience \cite{mahwish_measuring_nodate}. These traits serve as stable psychological anchors that interact dynamically with situational contexts in well-established fashions ~\cite{mejia_emotional_2015, ungar_resilience_2013, joiner_jr_depression_2009}.
    
    %
    
    

By combining these features, we constructed a baseline model that aligns with psychological theory, providing an interpretable reference point against which more data-driven approaches can be evaluated. Our method represents a true interdisciplinary effort in computational social science, bridging insights from personality psychology, emotion theory, and NLP to advance the study of mental health dynamics in digital contexts.

\subsection{Human Language Modeling: HaRT for Person-Contextual Embeddings}

Our principled baseline approach offers a clear explanatory mechanism for predicting well-being and distinguishing between adaptive and maladaptive self-states. However, we anticipate that more advanced language models will enhance predictive accuracy and provide a richer representation of language and individuals. \hart, trained on the Human Language Modeling (HuLM) task — which predicts the next word based on prior words, incorporating a latent user representation derived from their temporal historical language — enables a person-contextualized understanding of language ~\cite{soni-etal-2022-human}. Grounded in psychometric theory on the stability of psychological traits ~\cite{watson_stability_2004} HuLM processes an author's language collectively, recognizing that linguistic patterns are best understood within the context of the individual themselves, over time~\cite{soni2024proceedings, ganesan-etal-2024-text}. This approach is particularly well-suited for our tasks, given the dataset's longitudinal structure, where language is nested within individuals, and has proven to be effective in mental health assessments~\cite{v-ganesan-etal-2022-wwbp, varadarajan2024archetypes}, psychological assessments~\cite{soni-etal-2025-evaluation}, and user attributes assessments~\cite{soni-etal-2024-comparing}.




\section{Data \& Tasks}

\paragraph{Dataset.} 
The CLPsych 2025 shared task ~\cite{tseriotou2025overview} provided annotated evidence for adaptive and maladaptive self-states ~\cite{slonim2024self} as spans of texts from posts written by individuals historically in addition to a score representing the overall well-being in a post. The data consists of 30 users (timelines) with a total of 343 posts of which 199 posts were annotated. 

\paragraph{Shared Tasks.}
The shared tasks focus on the longitudinal modeling of changes in individual's mood and states  ~\cite{shing-etal-2018-expert, zirikly-etal-2019-clpsych, tsakalidis-etal-2022-overview}. We participate in 2 sub-tasks targeted at post-level judgments: a) predicting the overall well-being, and b) identifying evidence for adaptive and maladaptive self-states.

\section{Methods}

We extracted two categories of features: a theory-informed baseline — comprising Situational 8 DIAMONDS (S8D) and Person-Level Traits (PLT; see Section~\ref{sec: princi_base}) — and person-contextualized embeddings. While PLT features were computed at both sentence and post levels, S8D were limited to post-level annotations due to their reliance on broader context.

\paragraph{Situational 8 DIAMONDS (\situa).}
We used Deepseek-R1 \cite{deepseek-ai_deepseek-r1_2025} with few-shot prompting to infer scores for each of the eight situational dimensions at the post level. Each dimension was prompted separately using two manually annotated exemplars tailored to its psychological construct (see \S~\ref{apendix_situa}). Scores ranged from 1 (not present) to 9 (highly present), reflecting the inferred prominence of each situational characteristic.

\paragraph{Person-Level Traits (\plt).} We extracted 19 features across four subdomains:

\noindent\textit{Implicit Motives.} Following ~\citet{nilsson2025automatic}, we applied fine-tuned RoBERTa-Large models to estimate three subconscious motives — achievement, affiliation, and power — at the sentence level. These predictions were averaged and adjusted for word count to yield post-level scores (details in Appendix~\ref{apendix-implicit}).

\noindent\textit{Mental Health.} Using the Language-Based Assessment Model Library \cite{nilsson_language-based_nodate}, we inferred six psychological dimensions: valence \cite{eijsbroek_comparison_2023}, harmony in life, satisfaction with life~\cite{kjell2022natural}, anxiety, and two depression indices~\cite{gu2024natural}. Features were extracted at the sentence level and averaged to generate post-level estimates.

\noindent\textit{Resilience.} We implemented the Resilience through Language Modeling (\resili) framework \cite{mahwish_measuring_nodate} to compute scores for nine resilience-related facets (e.g., optimism, coping toolkit) at both sentence and post levels. See Appendix~\ref{apendix_resili} for details.

\noindent\textit{Cognitive Distortions.} Drawing on prior work~\cite{varadarajan-etal-2025-linking}, we used pretrained models to estimate levels of cognitive distortion, a known correlate of maladaptive emotional states~\cite{mann2002cognitive,bathina2021individuals}, at both sentence and post levels.



\paragraph{Person-Contextual Embeddings: HaRT.}
We fine-tuned HaRT (Human-aware Recurrent Transformer)~\cite{soni-etal-2022-human} to predict continuous well-being scores at the post level and binary adaptive/maladaptive labels at the sentence level. To do so, we split the CLPsych training data into internal training and validation sets. HaRT processes users' historical posts in sequence, enabling the generation of temporally informed, person-specific embeddings at both the sentence and post levels.

We evaluated these embeddings across three tasks using 5-fold nested cross-validation: (a) continuous well-being prediction via ridge regression, (b) adaptive label prediction using logistic regression, and (c) maladaptive label prediction using logistic regression. We chose 5-fold CV to mitigate overfitting given the small sample size. For all classifiers, we used a penalty range of [10, 0, -1, -0.10, 0.10]. For span identification, we predicted label probabilities and applied thresholds of 0.45 (adaptive) and 0.4 (maladaptive) to extract evidence-level annotations.

\section{Results \& Discussion}

\begin{table}
    \centering
    \begin{small}
    \begin{tabular}{lcc}
        \toprule
        & \multicolumn{2}{c}{\textbf{5-fold Ridge CV}} \\
        & $r$ ↑ & $MSE$ ↓  \\ 
        \midrule
        \situa & 0.528 & 2.556 \\
        \disto & 0.365 & 3.059 \\
        \resili & 0.533 & 2.538 \\
        \plt & \textbf{0.629} & \textbf{2.149} \\
        \situa + \resili + \disto & 0.623 & 2.178 \\
        \situa + \plt  &0.622  & 2.174 \\
        \bottomrule
    \end{tabular}
    \end{small}
    \caption{Pearson correlation (r) and Mean Squared Error (MSE) when training a ridge regression model using different ``principled'' baseline features to predict continuous Well-being scores using nested 5-fold cross-validation.}
    \label{tab:internal_wellbeing_CV}
\end{table}

\begin{table}
    \centering
    \begin{small}
    \begin{tabular}{lcc | cc}
        \toprule
        & \multicolumn{2}{c}{\thead{\textbf{Internal} \\ \textbf{Val Set}}} & \multicolumn{2}{c}{\thead{\textbf{5-fold} \\ \textbf{Ridge CV}}} \\
        & $r$ ↑ & $MSE$ ↓ & $r$ ↑ & $MSE$ ↓  \\ 
        \midrule
        \thead[l]{\hartwbft} & 0.684 & 1.828 & 0.876 & 0.828 \\
        \thead[l]{\hartwbft + \situa \\ + \resili + \disto} & - & - &  0.883& 0.787 \\
        \thead[l]{\hartwbft + \\ \situa + \plt} & - & - & 0.884 & 0.783 \\
        \bottomrule
    \end{tabular}
    \end{small}
    \caption{Pearson correlation (r) and Mean Squared Error (MSE) when fine-tuning \hart using internal train and validation splits and further training a ridge regression model using resulting embeddings and principled baseline features. Note: Pearson $r$ values for the 5-fold ridge CV numbers are likely inflated due to partial data contamination across the fine-tuning and cross-validation datasets. The internal validation set was used while finetuning the HaRT model, after which the weights were separately used as inputs for well-being task. The internal validation numbers have been omitted for the post-finetuned model.}
    \label{tab:internal_wellbeing_Val}
\end{table}

\paragraph{Well-being Scores.}
Situational characteristics (\situa) inferred from posts were predictive of annotated well-being scores (see Table~\ref{tab:internal_wellbeing_CV}). When combined with PLT features, our “principled” baseline — grounded in interactionist theory — yielded improved performance. The psychometric theory-inspired \hart model outperformed baselines on the internal validation set, although we observed signs of overfitting in 5-fold CV (see Table~\ref{tab:internal_wellbeing_Val}). Nonetheless, official results showed similar trends (Table~\ref{tab:official_wellbeing}), with the theory-driven \situa+\plt baseline outperforming the theory-agnostic \hartwbftridge variant.

\begin{figure}
    \centering
    \begin{minipage}{0.49\textwidth}
        \includegraphics[width=0.95\linewidth]{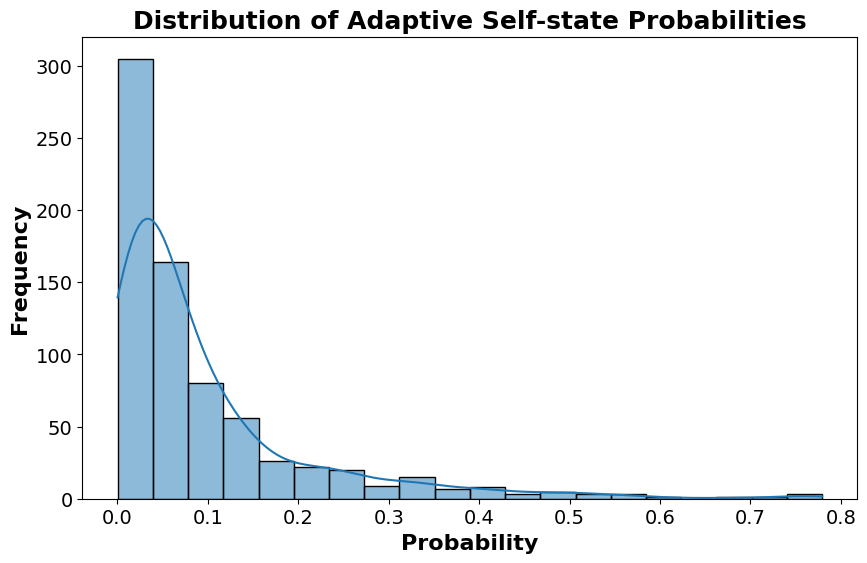}
    \end{minipage}
    \begin{minipage}{0.49\textwidth}
        \includegraphics[width=0.95\linewidth]{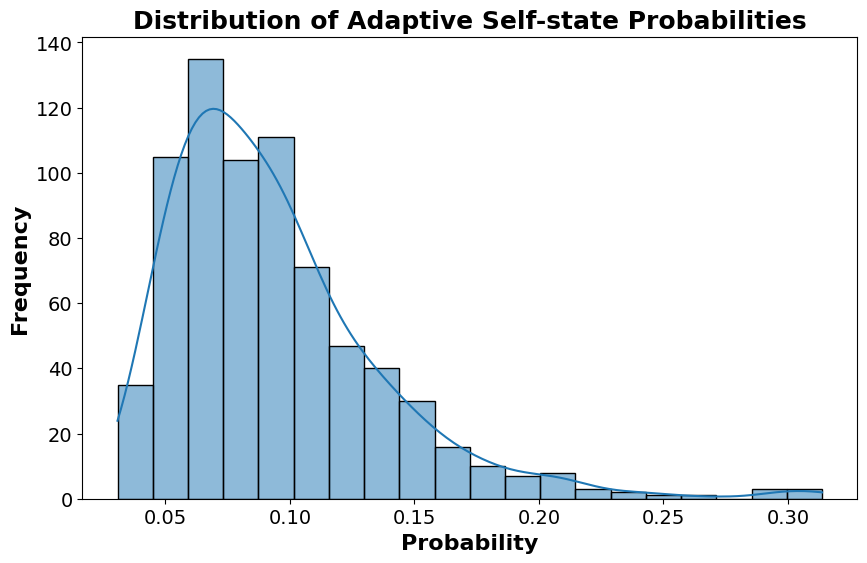}
    \end{minipage}
    \caption{Distribution of probabilities to predict adaptive state for a given sentence. On the top is using \hartwbft + PLT features, and the bottom is using PLT features in Logistic Regression models.}
    \label{fig:adap_probs}
\end{figure}

\paragraph{Adaptive and Maladaptive States.}
\hart models performed well in the binary classification of adaptive and maladaptive self-states (Table~\ref{tab:internal_adap_maladap}), with additional gains observed when combined with PLT features (Table~\ref{tab:internal_adap_maladap_CV}). While \disto and \plt features alone showed reasonable performance, they produced minimal variation in predicted probabilities across examples (Figure~\ref{fig:adap_probs}). In contrast, \hart-based models exhibited greater sensitivity to language variation and were more effective at identifying adaptive and maladaptive evidence spans. Additional supporting results and probability distributions can be found in Appendix \S~\ref{app:supp_results}.

\begin{table}
\small
    \centering
    \begin{small}
    \begin{tabular}{lccc}
        \toprule
        & \multicolumn{2}{c}{\textbf{5-fold Log.reg. CV}} \\
        & $F1_{macro}$ & $AUC$   \\ 
        \midrule
         \disto\tiny{adaptive}  & \textbf{0.54}  & 0.75 \\
        \hartwbft & 0.50  & 0.74\\
        \hartwbft + \resili + \disto\tiny{adaptive}  & 0.53 & \textbf{0.76}\\
        \plt\tiny{adaptive}  & 0.48 & 0.66 \\
        \hartwbft + PLT\tiny{adaptive} & 0.52 & \textbf{0.76} \\
        \midrule
         \disto\tiny{maladaptive}  & 0.56  & 0.73 \\

         
        \hartwbft  & 0.56  & 0.73 \\
        \hartwbft + \resili + \disto\tiny{maladaptive}  &  0.57 &0.76 \\

        \plt\tiny{maladaptive}  & 0.49 & 0.70 \\

         \hartwbft + PLT\tiny{maladaptive}  & \textbf{0.58} & \textbf{0.77} \\
         \bottomrule
    \end{tabular}
    \end{small}
    \caption{Macro F1 and AUC results when training a logistic regression model to predict binary adaptive and maladaptive labels separately over sentences split from posts.}
    \label{tab:internal_adap_maladap_CV}
\end{table}






\subsection{Discussion.} 
\begin{figure}[t]
    \centering
    \begin{minipage}{0.49\textwidth}
        \includegraphics[width=\linewidth, height=6cm]{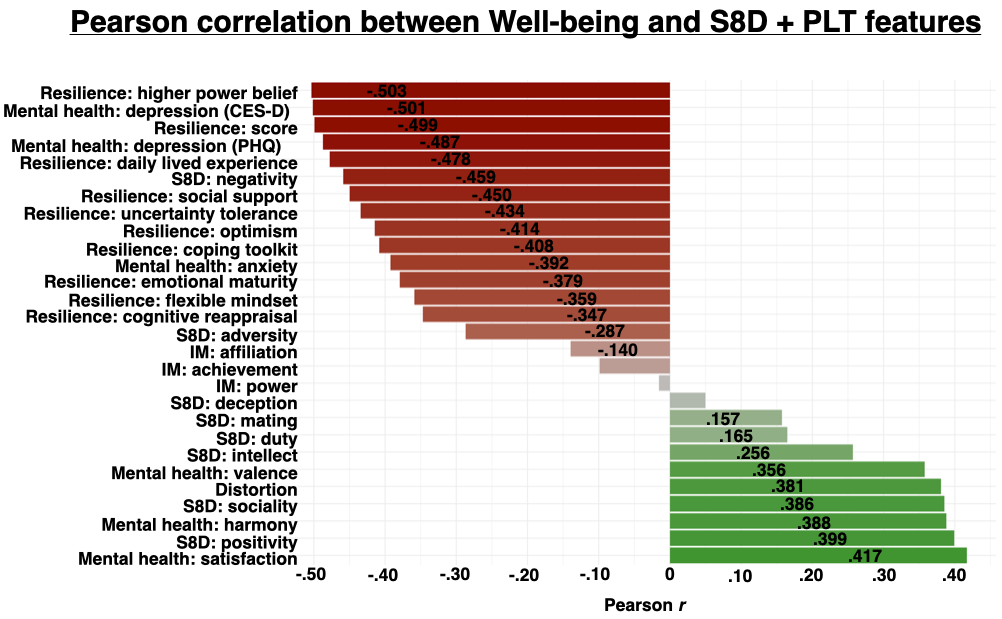}
    \end{minipage}
    \hfill
    \begin{minipage}{0.49\textwidth}
        \includegraphics[width=\linewidth, height=6cm]{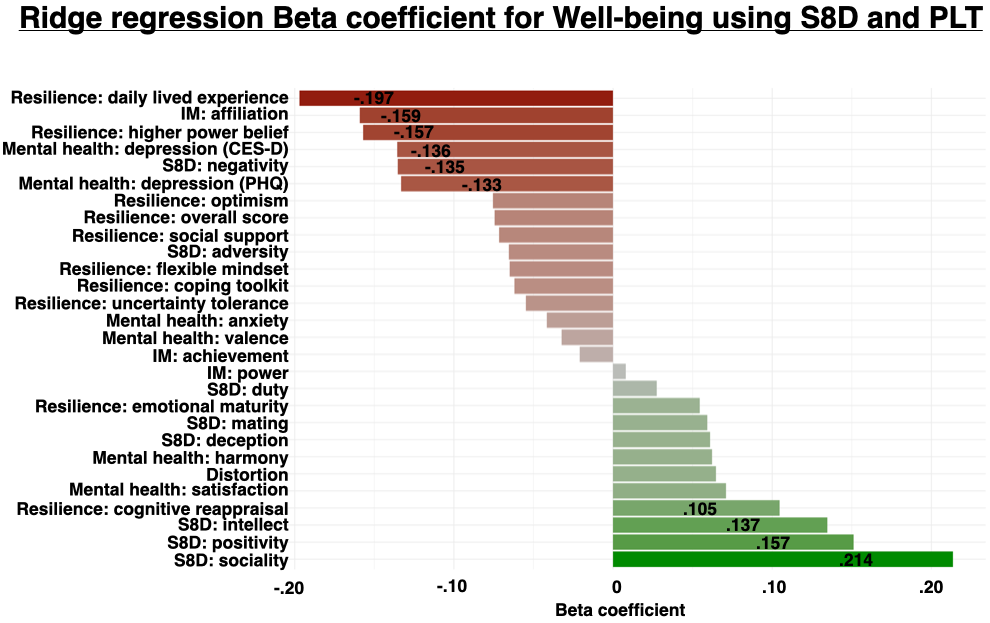}
    \end{minipage}
    \caption{Qualitative analysis of features in our principled baseline consisting of psychological characteristics of the situation and person-level traits. Left: Pearson correlation coefficients; Right: Ridge regression beta coefficients for predicting well-being with the \situa and \plt features.}
    \label{fig:pearson_beta_wb_plt_s8d}
\end{figure}

Our interactionist, theory-based ``principled'' baseline approach effectively predicts annotated well-being scores. However, it struggled to capture evidence of adaptive and maladaptive states within posts, highlighting the challenge of disentangling self-states from their situational context — an issue that even human observers can find difficult to assess accurately ~\cite{uleman_people_1996, nisbett_behavior_1973, ross_intuitive_1977}.

To further explore the predictive power of our principled baseline features, we conduct a qualitative analysis of well-being correlations. As shown in Figure \ref{fig:pearson_beta_wb_plt_s8d}, the top three features positively associated with well-being scores are: `satisfaction with life' (from mental health in PLT), `positivity' in the situation (from \situa), and `harmony in life' (from mental health in PLT). Conversely, the top three features negatively correlated with well-being scores include: `higher power belief' (resilience from PLT), `depression scale' (mental health from PLT), and overall `resilience score' (from PLT).

Additionally, Figure \ref{fig:pearson_beta_wb_plt_s8d} shows that the ridge regression model, leveraging our principled baseline features, assigns positive importance to `sociality', `positivity', and `intellect' (all from \situa), while attributing negative importance to `daily lived experience' (from resilience in \plt), `need for affiliation' (from implicit motives in \plt)\footnote{High expression of affiliation-related language may indicate a frustrated, rather than fulfilled, need for social connection. Classic motivation theories \citep{mcclelland_human_1987} suggest that individuals who frequently verbalize their affiliation motive may be experiencing social deprivation or unmet interpersonal needs. This aligns with research on compensatory behaviors, where socially disconnected individuals often amplify affiliative overtures to seek connection \citep{baumeister_need_1995,richman_reactions_2009}.}, and `belief in a higher power' (from resilience in \plt).

These findings align with prior research indicating that well-being is closely tied to life satisfaction and positive social interactions ~\cite{diener_subjective_1999, cacioppo_social_2014}. Similarly, negative associations with depression, present-focus, and certain aspects of religiosity are consistent with existing psychological literature on mental health dynamics ~\cite{beck_depression_1967, himmelstein_linguistic_2018, mccullough_religion_1999, braam_religion_2019}. 
A more detailed discussion of these results can be found in the Appendix \S~\ref{app:cont_disc}.




\section{Conclusion}

Mental health is not a static trait but a dynamic outcome shaped by ongoing interactions between person and context. In this work, we operationalized this psychological insight by combining person-level traits and situational characteristics — core tenets of interactionist and constructionist theory — to model well-being and adaptive self-states in language.

Our theory-driven baseline, built from the Situational 8 DIAMONDS and language-inferred psychological traits, demonstrated strong performance while offering interpretable, psychologically grounded predictions. Features like positivity, satisfaction with life, and harmony in life emerged as key indicators of well-being, while markers of cognitive distortion and unmet affiliation needs were linked to maladaptive patterns. HaRT’s person-contextualized embeddings added value in modeling temporal variation, particularly for adaptive and maladaptive evidence detection.

These findings highlight the value of bridging computational models with psychological theory — not only to improve prediction, but to ensure outputs are meaningful and human-understandable. Future work should explore how different contexts modulate trait adaptiveness, and how language-based systems might support more flexible, resilient self-states over time.

By integrating theory and computation, we move toward systems that understand individuals not as fixed entities, but as contextually situated and dynamically evolving.

\section*{Limitations}

While this work offers a psychologically grounded and interpretable approach to modeling mental health from language, several limitations must be acknowledged — both technical and conceptual.

First, our analyses are constrained by the scale and structure of the CLPsych 2025 dataset. With only 30 users and fewer than 200 annotated posts, the generalizability of our findings is limited. Although we used robust cross-validation and avoided tuning on the test set, future work should evaluate these models on larger, more diverse, and demographically representative datasets.

Second, the ground truth labels themselves are inherently interpretive, reflecting human judgments of well-being and self-states based on textual evidence. This raises a broader epistemological question: to what extent can self-states be reliably inferred from language alone? Our work assumes that linguistic expressions are sufficient proxies for psychological states — a premise that, while useful for modeling, must be critically examined in clinical and applied settings.

Third, our feature extraction relies on pretrained models and heuristics that may carry latent biases or be insensitive to cultural or contextual nuance. For example, expressions of distress or resilience may vary across communities, and models trained on general corpora may fail to capture such variation meaningfully.

Finally, although we draw from psychological theory, our models remain correlational. They can identify linguistic markers of mental health but in no way are able to definitively speak to underlying mechanisms, causal relationships, or interventions. Future work should incorporate longitudinal clinical assessments to validate language-based features against real-world outcomes.

These limitations do not undermine the value of this work, but rather highlight the need for computational psychology to remain grounded in balancing predictive power with interpretive care, and data-driven modeling with theoretical accountability.

\section*{Ethical Considerations}

Modeling mental health from language data presents profound ethical challenges. While the tools developed in this work aim to advance understanding of mental states in contextually sensitive and interpretable ways, their misuse — or even well-intentioned use without adequate safeguards — poses real risks to privacy, autonomy, and well-being.

First and foremost is the issue of consent. Although the data used in this study were shared with participant permission as part of a structured research challenge, this controlled environment does not reflect broader real-world settings in which language-based models might be applied. Any future deployment must ensure that individuals are aware of — and have control over — how their language data are interpreted, stored, and acted upon.

Second, language-based inferences about mental health are probabilistic and inherently contain some degree of uncertainty. Over-reliance on model outputs — particularly in clinical, legal, or surveillance contexts — could lead to misdiagnoses, stigmatization, or unwarranted interventions. The interpretability of our features helps mitigate this risk, but human oversight and psychological expertise remain essential in any applied use.

Third, there are critical concerns around representation and bias. Our models are trained on English-language data from social media forums, which may reflect particular cultural, demographic, and socioeconomic perspectives. As a result, model outputs may not generalize across populations and could even reinforce existing inequities if deployed indiscriminately. Expanding the diversity of training data and engaging with cultural psychology are necessary steps forward.

Finally, as researchers in computational social science, we must remain vigilant about the institutional and commercial pressures that can shape how mental health technologies are built and used. The potential to infer mental states from language at scale invites both promise and peril. Ethical research in this space demands more than compliance — it requires an ongoing commitment to transparency, self-reflection, and the prioritization of human dignity.

\section*{Acknowledgments}
We are grateful to the organizers of the CLPsych 2025 Shared Task for their efforts in curating the dataset, designing an impactful task, and fostering interdisciplinary collaboration. \\
This research is supported in part by the Office of the Director of National Intelligence (ODNI), Intelligence Advanced Research Projects Activity (IARPA), via the HIATUS Program contract \#2022-22072200005, a grant from the NIH-NIAAA (R01 AA028032), and a grant from the CDC/NIOSH (U01 OH012476), Developing and Evaluating Artificial Intelligence-based Longitudinal Assessments of PTSD in 9/11 Responders. The views and conclusions contained herein are those of the authors and should not be interpreted as necessarily representing the official policies, either expressed or implied, of ODNI, IARPA, any other government organization, or the U.S. Government. The U.S. Government is authorized to reproduce and distribute reprints for governmental purposes notwithstanding any copyright annotation therein.

\bibliography{custom, ryanboyd}
\setcounter{table}{0}
\renewcommand{\thetable}{A\arabic{table}}
\setcounter{figure}{0}
\renewcommand{\thefigure}{A\arabic{figure}}

\appendix

\section{Appendix}
\label{sec:appendix}

\subsection{Official Submissions and results}

Table~\ref{tab:official_adap_maladap} shows the official shared task results for Task A.1 (evidence extraction), and Table~\ref{tab:official_wellbeing} shows the official results for Task A.2 (wellbeing prediction).

\begin{table*}
    \centering
    \begin{tabular}{lccc | ccc}
        \toprule
        & \multicolumn{6}{c}{\textbf{A.1 Adaptive/Maladaptive}} \\
        & \multicolumn{3}{c}{\textbf{recall}} & \multicolumn{3}{c}{\textbf{weighted recall}} \\
        & \thead{overall} & \thead{adaptive} & \thead{maladaptive} 
        & \thead{overall} & \thead{adaptive} & \thead{maladaptive} \\
        \midrule

        \hartwbft + \resili + \disto (LinSVC) & 0.108 & 0.099 & 0.116 & 0.103 & 0.097 & 0.109 \\
        \hartwbft \tiny{+ Log.reg.} & 0.105	& 0.077	& 0.132	& 0.102	& 0.075	& 0.129 \\
        \hartadapft & \textbf{0.276}	&\textbf{ 0.245}	& \textbf{0.308	}& - & - & - \\
        \hartmaladapft & - & - & -	& \textbf{0.236}	& \textbf{0.238}	& \textbf{0.235 }\\
        
        \bottomrule
    \end{tabular}
    \caption{Task A.1 Adaptive and Maladaptive Evidence task. We found that finetuning HaRT for independent adaptive and maladaptive evidence classification can yield significant boosts over traditional principled baselines.}
    \label{tab:official_adap_maladap}
\end{table*}

\begin{table*}
    \centering
    \begin{tabular}{lcccc | c}
        \toprule
        & \multicolumn{5}{c}{\textbf{A.2 Well-being}} \\
        & \multicolumn{4}{c}{\textbf{timeline MSE} ↓} & \textbf{F1} \\
        & \thead{overall} & \thead{minimal \\ impairment} & \thead{impaired} & \thead{serious \\ impairment} & \thead{macro}  \\
        \midrule

        \hartwbft & 3.73	& 2.76	& 1.95	& 5.92	& 0.15 \\
        \hartwbftridge & 3.42	& 2.95	& \textbf{1.33}	& 5.1	& 0.17 \\
        \hartwbft + \situa + \resili + \disto (Ridge) &	3.27 &	2.63 &	1.38 &	4.98 &	0.19 \\
        \hartwbft + \situa + \plt (Ridge) & 3.22	& 1.40	& 2.60	& 4.86 & 0.19 \\
        \situa + \resili + \disto (Ridge) & 3.02	& \textbf{1.21}	& 1.79	& 5.25 & 0.17 \\
        \situa + \plt (Ridge) & \textbf{2.78}	& 1.84	& 2.14	&\textbf{ 3.89} & 0.19 \\

        \bottomrule
    \end{tabular}
    \caption{Task A.2 Wellbeing task. We found that unlike Task A.1, finetuning HaRT to the wellbeing prediction task need not consistently offer boosts, instead, principled and theoretical methods can offer significant advantages, with a small number of interpretable dimensions without compromising on the accuracy. The first 3 rows in this table were our official submissions while others are presented for additional analysis. We note that using HaRt fine-tuned for the respective adaptive and maladaptive binary classifications may provide benefits over using \hartwbft in all combinations, however, due to time constraints we do not have empirical results for the same.}
    \label{tab:official_wellbeing}
\end{table*}

\subsection{Situational 8 DIAMONDS Prompts}
\label{apendix_situa}

We used the items associated with each psychosocial situation from \citet{rauthmann_situational_2014} to define each of the S8D to Deepseek-R1. We individually prompted for each of the S8D and provided two personalized exemplars and annotations for each of the exemplars (2-shot). Some few-shot examples were manually created and some were picked from the CLPsych data. The prompts template would read as follows:

\begin{quote}
\begin{itshape}
Instruction: You are an expert in situational perception and psychological analysis. Your task is to evaluate a given block of text for the (insert situation) dimension from the Situational 8 DIAMONDS taxonomy. Individuals who score higher in the (insert situation) dimension relate to the following situations:

\{\{S8D items corresponding to relevant dimension were inserted here\}\}

Your task is to provide the following in a structured JSON format: Rating: Assign a numerical rating for the mating dimension on a scale of 1 to 9 (where 1 = Not at all present and 9 = Highly present). 

Reasoning: Provide a justification for the rating based on the text. Span Extraction: Identify specific phrases in the text that support your rating. 

Below are two examples with respective input texts and corresponding outputs to illustrate the task: 

\{\{Example texts with corresponding annotations were inserted here\}\}

Now, evaluate the following input text:

\{\{Text requiring annotation was inserted here\}\}
\end{itshape}
\end{quote}

We then curated two examples catered towards each situation and hand-annotated the example to provide as a guideline for output. We provided two messages from the CLPsych 2025 dataset for the situations Duty and Intellect, although for the remaining six S8D, we curated our own examples. An example annotation for the situation Adversity (including manually creates few-shot examples) goes as follows: 

\begin{quote}
"Example 1: "I can’t believe how unfair my manager is being. I worked overtime all last week, skipped my breaks, and still got blamed for a project delay that wasn’t even my fault. He called me out in front of the entire team, making it sound like I was slacking off. I tried to defend myself, but he just dismissed me and moved on. It’s exhausting constantly feeling like I have to prove myself, only to be treated like I’m incompetent." 

**Output:** "adversity": 8, "reasoning": "The individual is experiencing direct blame from their manager for a project delay that was not their fault. They describe being publicly criticized in front of colleagues and dismissed when attempting to defend themselves. The tone reflects frustration and exhaustion from repeated unfair treatment, which strongly aligns with the Adversity dimension.","supporting spans": "I worked overtime all last week, skipped my breaks, and still got blamed for a project delay that wasn’t even my fault.","He called me out in front of the entire team, making it sound like I was slacking off.","I tried to defend myself, but he just dismissed me and moved on.","It’s exhausting constantly feeling like I have to prove myself, only to be treated like I’m incompetent." 

Example 2: "Every time I try to express my opinion, my older brother just shuts me down. He talks over me, mocks what I say, and makes me feel like I’m too stupid to contribute. It’s like my thoughts don’t matter in my own family. Even when I call him out on it, he just laughs and says I’m being too sensitive. I don’t know how to get him to take me seriously."

**Output:** "adversity": 7, "reasoning": "The individual describes repeated experiences of being dismissed, mocked, and dominated by their older brother. The situation involves verbal criticism, a power imbalance, and an inability to be taken seriously, all of which strongly align with the Adversity dimension.", "supporting spans": "Every time I try to express my opinion, my older brother just shuts me down.", "He talks over me, mocks what I say, and makes me feel like I’m too stupid to contribute.", "Even when I call him out on it, he just laughs and says I’m being too sensitive.""

\end{quote}

\subsection{Implicit Motives and Mental Health in PLT}
\label{apendix-implicit}

To construct our person-level trait (PLT) features, we extracted both implicit motivational needs and core mental health dimensions from participants’ language using pre-existing, validated models.

\paragraph{Implicit Motives.}  
Following classic motivational theory \cite{mcclelland_human_1987}, we define three core implicit motives reflected in language: (1) the need for achievement, indicated by references to striving for excellence; (2) the need for affiliation, reflected in efforts to initiate or maintain friendly relationships; and (3) the need for power, expressed as influence or control over others or institutions. We used RoBERTa-based models from prior work~\cite{nilsson2025automatic}, trained on expert-coded Picture Story Exercises, to infer these motives at the sentence level. Sentence-level predictions were then aggregated to post level using word count-adjusted averaging procedures.

\paragraph{Mental Health Dimensions.}  
We further extracted six features representing key aspects of mental health using models from the Language-Based Assessment Model Library \cite{nilsson_language-based_nodate}. These include:

\begin{itemize}
    \item \textbf{Valence:} Trained on annotated Facebook posts rated for emotional positivity or negativity. The model’s out-of-sample correlation with human ratings was $r = .81$.
    \item \textbf{Harmony in Life \& Satisfaction with Life:} Trained on open-text responses rated using validated scales \cite{kjell_natural_2022}. The models achieved out-of-sample correlations of $r = .73$ and $r = .71$, respectively.
    \item \textbf{Depression:} Two separate models were used — one trained to the Patient Health Questionnaire-9 (PHQ-9; \citealp{kroenke_patient_2011}), and the other to the Center for Epidemiologic Studies Depression Scale (CES-D; \citealp{radloff_ces-d_1977}). These models yielded correlations of $r = .66$ and $r = .73$, respectively.
    \item \textbf{Anxiety:} Trained on worry-based language mapped to the Generalized Anxiety Disorder 7-item scale (GAD-7), with a correlation of $r = .63$.
\end{itemize}

All models were previously pre-registered for their respective source projects, evaluated using nested cross-validation, and applied out-of-sample to the present dataset. These features form part of our psychologically interpretable PLT baseline.

\subsection{Resilience in PLT}
\label{apendix_resili}

Traditional views of resilience often reduce it to the absence of psychopathology or the ability to recover from stress. However, contemporary psychological frameworks emphasize a broader understanding: resilience as a multidimensional capacity for adaptive functioning in the face of adversity. To operationalize this richer perspective, we used the ReLM (Resilience using Language Modeling) framework \cite{mahwish_measuring_nodate}, which integrates an archetype-based approach to assess resilience from language.

ReLM captures nine core facets of resilience: optimism, flexibility mindset, sense of social support (SoS), continued activities of daily living (CADL), cognitive reappraisal, emotional maturity, uncertainty tolerance, belief in a higher power, and coping toolkit. Each facet is represented by four prototype statements — brief exemplar sentences derived from a synthesis of resilience literature and analysis of archival interviews with individuals who have demonstrated stability in the face of trauma. For example, a prototype for flexibility mindset reads: “I always try new things because I'm open to exploring.”

To assess individual alignment with each facet, ReLM embeds both prototype statements and participant text using Sentence RoBERTa and computes their semantic similarity \cite[see:][]{varadarajan2024archetypes}. The resulting scores quantify how strongly a participant's language reflects each dimension of resilience. A higher score indicates greater expression of that facet.

Finally, a composite resilience score is computed by applying exploratory factor analysis to the nine facet scores. Across multiple datasets, a single-factor solution consistently explained 49–56\% of the variance \cite{mahwish_measuring_nodate}, supporting the use of a unified resilience metric. This composite score provides a theoretically grounded and interpretable estimate of an individual's language-based resilience profile.

\subsection{Continued Discussion}
\label{app:cont_disc}
The `daily lived experience' facet in this dataset exhibits a strong negative association with well-being. This relationship likely stems from the facet’s focus on individuals persisting through routine tasks despite ongoing stressors. Participant statements such as “I’ve adjusted to my new environment—it was hard at first, but I’m improving” and “I’ve been leaving the house more” (rewritten for anonymity) illustrate gradual adaptation and effort. However, because well-being in this dataset is framed in terms of symptom absence and unimpaired functioning, the `daily lived experience' facet presents a paradox. While it reflects resilience — people continuing daily tasks despite struggles — it also signals underlying difficulty. The very act of pushing forward in the face of these challenges may indicate diminished well-being, as it suggests persistent symptomatology masked by forced functionality.

The Belief in a Higher Power facet reflects trust in external forces during times of struggle, as illustrated by participant statements like “I accept things as they are… I trust that things will get better over time” and “I know that life will work itself out” (anonymized). These responses suggest a surrender of personal control to fate or higher being—a mechanism that may offer emotional relief. However, the model’s negative weighting of this facet stems from a tension between its definition of well-being (rooted in agency, engagement, and lack of symptom) and a resilience strategy that relies on external control. While faith can provide comfort, the framework may be interpreting passive reliance on higher powers as maladaptive in contexts where well-being is tied to active mastery of one’s circumstances.

\subsection{Supplementary results and figures}
\label{app:supp_results}

\begin{table*}
\small
    \centering
    \begin{small}
    \begin{tabular}{lcccc}
        \toprule
        & \multicolumn{4}{c}{\textbf{Internal Val Set}} \\
        & $F1_{macro}$ & $F1_{wtd}$ & $AUC$ & $Acc$ \\ 
        \midrule
         \disto\tiny{adaptive} & 0.53 & 0.85 & 0.75 & 0.89  \\
        \hartadapft & 0.67 & 0.89 & 0.66 & 0.89 \\
         \disto\tiny{maladaptive} & 0.56 & 0.82 & 0.76 & 0.84 \\

         \hartmaladapft & 0.60 & 0.82 & 0.60 & 0.82 \\
         
         \bottomrule
    \end{tabular}
    \end{small}
    \caption{Task A.1 Additional results on internal validation set when predicting binary adaptive and maladaptive labels separately over sentences split from posts.}
    \label{tab:internal_adap_maladap}
\end{table*}

\begin{table*}
\small
    \centering
    \begin{small}
    \begin{tabular}{lcccc}
        \toprule
        & \multicolumn{4}{c}{\textbf{5-fold Log.reg. CV}} \\
        & $F1_{macro}$ & $F1_{wtd}$ & $AUC$ & $Acc$  \\ 
        \midrule
         \disto\tiny{adaptive}  & 0.54 & 0.87 & 0.75 & 0.90 \\
        \hartwbft & 0.50 & 0.87 & 0.74& 0.90\\
        \hartwbft + \resili + \disto\tiny{adaptive}  & 0.53 & 0.87 & 0.76 & 0.91\\
        \plt\tiny{adaptive}  & 0.48& 0.86 &0.66 & 0.91\\
        \hartwbft + PLT\tiny{adaptive} & 0.52 & 0.87 & 0.76 & 0.90\\
        \midrule
         \disto\tiny{maladaptive}  & 0.56 & 0.81 & 0.73 & 0.85 \\

         
        \hartwbft  & 0.56 & 0.81 & 0.73 & 0.85\\
        \hartwbft + \resili + \disto\tiny{maladaptive}  &  0.57 & 0.82 &0.76 & 0.86 \\

        \plt\tiny{maladaptive}  & 0.49& 0.80 & 0.70 & 0.86\\

         \hartwbft + PLT\tiny{maladaptive}  & 0.58& 0.83& 0.77& 0.86\\
         \bottomrule
    \end{tabular}
    \end{small}
    \caption{Additional results when training a logistic regression model to predict binary adaptive and maladaptive labels separately over sentences split from posts.}
    \label{tab:internal_adap_maladap}
\end{table*}

\begin{figure}
    \centering
    \begin{minipage}{0.45\textwidth}
        \includegraphics[width=0.95\linewidth]{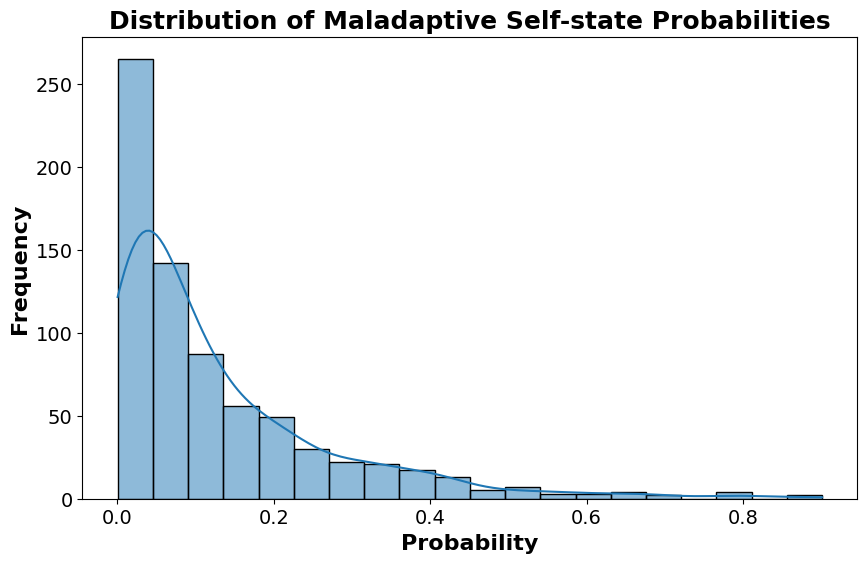}
    \end{minipage}
    \begin{minipage}{0.45\textwidth}
        \includegraphics[width=0.95\linewidth]{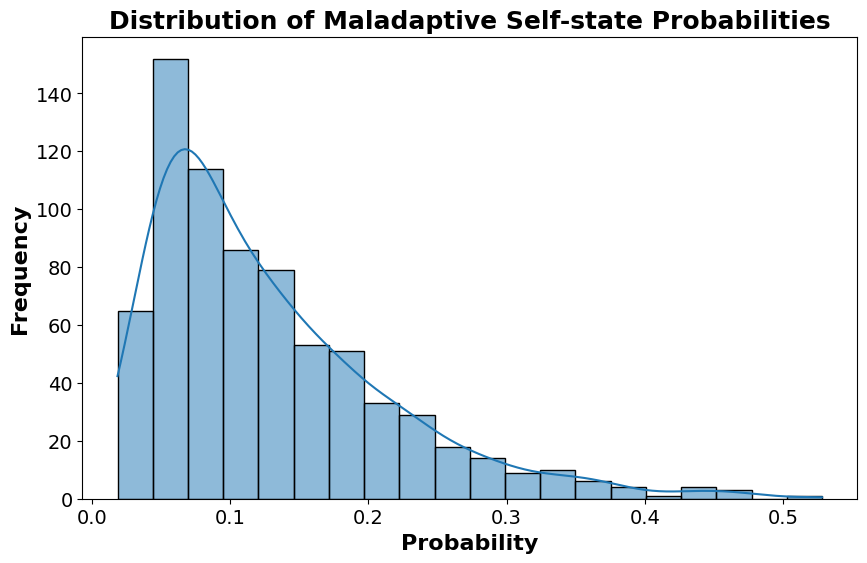}
    \end{minipage}
    \caption{Distribution of probabilities to predict maladaptive state for a given sentence. On the top is using \hartwbft + \plt features, and the bottom is using \plt features in Logistic Regression models.}
    \label{fig:maladap_probs_suppl_plt}
\end{figure}

\begin{figure}
    \centering
    \begin{minipage}{0.45\textwidth}
        \includegraphics[width=0.95\linewidth]{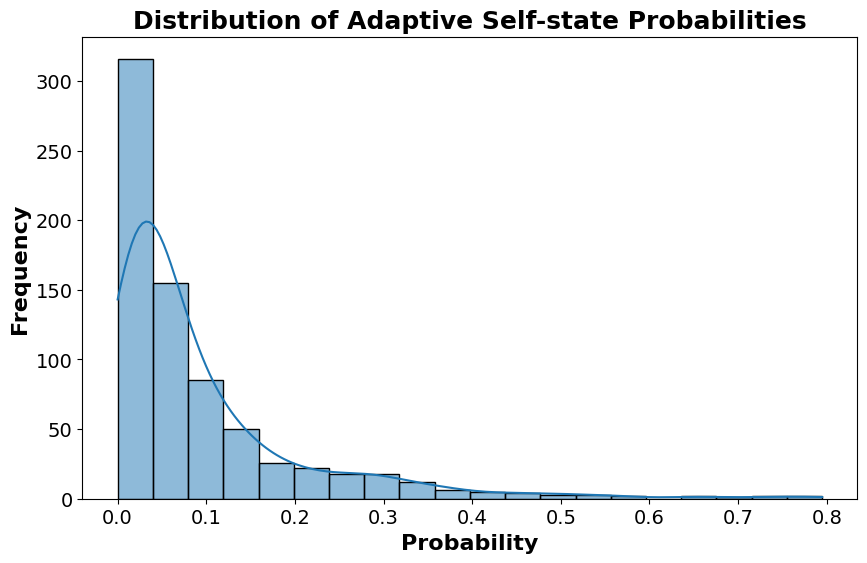}
    \end{minipage}
    \begin{minipage}{0.45\textwidth}
        \includegraphics[width=0.95\linewidth]{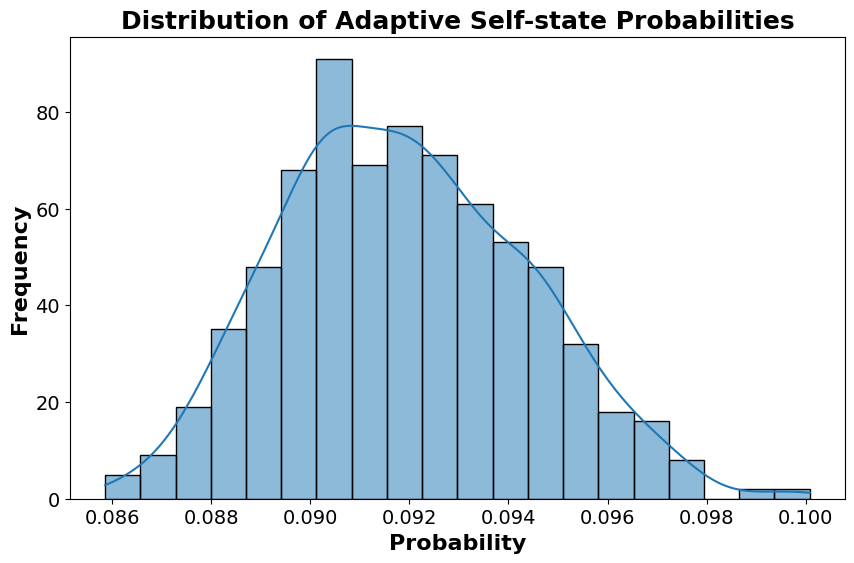}
    \end{minipage}
    \caption{Distribution of probabilities to predicting adaptive state for a given sentence. On the top is using \hartwbft + \resili + {\disto\tiny{adaptive}} features, and the bottom is using \resili + {\disto\tiny{adaptive}} features in Logistic Regression models.}
    \label{fig:adap_probs_suppl_resil_dist}
\end{figure}

\begin{figure}
    \centering
    \begin{minipage}{0.45\textwidth}
        \includegraphics[width=0.95\linewidth]{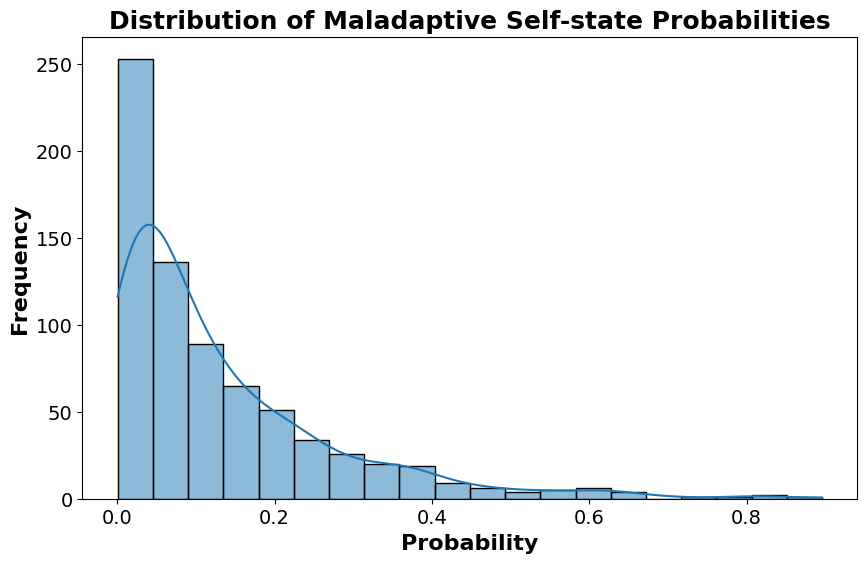}
    \end{minipage}
    \begin{minipage}{0.45\textwidth}
        \includegraphics[width=0.95\linewidth]{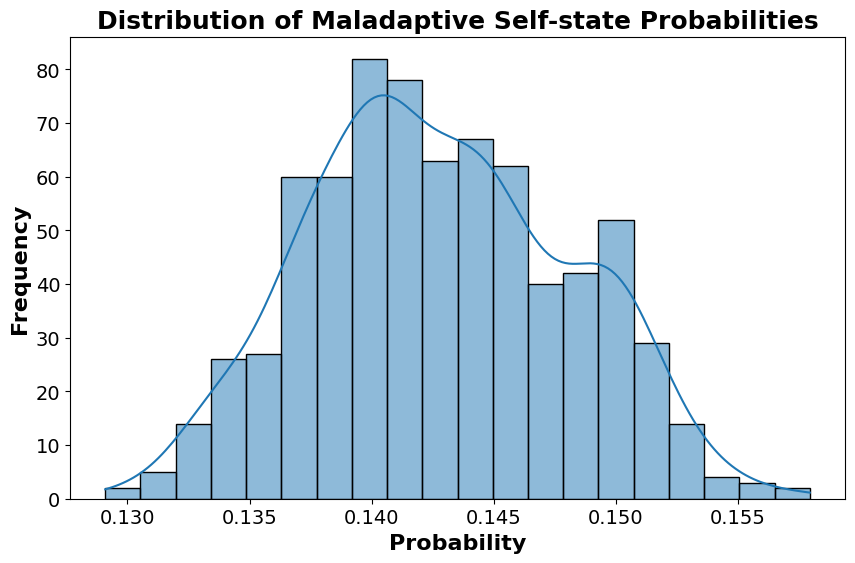}
    \end{minipage}
    \caption{Distribution of probabilities to predicting maladaptive state for a given sentence. On the top is using \hartwbft + \resili + {\disto\tiny{maladaptive}} features, and the bottom is using \resili + {\disto\tiny{maladaptive}} features in Logistic Regression models.}
    \label{fig:maladap_probs_suppl_resil_dist}
\end{figure}

\end{document}